\pdfoutput=1

\documentclass[11pt]{article}

\usepackage{EMNLP2022}

\usepackage{times}
\usepackage{latexsym}

\usepackage[T1]{fontenc}

\usepackage[utf8]{inputenc}

\usepackage{microtype}

\usepackage{inconsolata}

\usepackage{booktabs}
\usepackage{multirow}
\usepackage{graphicx}
\usepackage{subcaption}
\usepackage{amsmath}

\DeclareMathOperator{\uniform}{U}
\DeclareMathOperator{\binomial}{B}

\newcommand{\nmask}{n_\mathrm{mask}}
\newcommand{\nbidir}{n_\mathrm{bidir}}
\newcommand{\rbidir}{r_\mathrm{bidir}}
\newcommand{\npredict}{n_\mathrm{predict}}

\newcommand{\clm}{\textsc{NxtUni}}
\newcommand{\clmp}{\textsc{NxtPre}}
\newcommand{\mlm}{\textsc{MskBi}}
\newcommand{\mlmc}{\textsc{MskUni}}
\newcommand{\cmlm}{\textsc{HybUni}}
\newcommand{\cmlmp}{\textsc{HybPre}}

\title{On the Role of Bidirectionality in Language Model Pre-Training}

\author{Mikel Artetxe \quad Jingfei Du \quad Naman Goyal \quad Luke Zettlemoyer \quad Ves Stoyanov  \\
Meta AI \\
\texttt{\{artetxe,jingfeidu,naman,lsz,ves\}@meta.com}
}

\begin{document}
\maketitle
\begin{abstract}
Prior work on language model pre-training has explored different architectures and learning objectives, but differences in data, hyperparameters and evaluation make a principled comparison difficult. In this work, we focus on bidirectionality as a key factor that differentiates existing approaches, and present a comprehensive study of its role in next token prediction, text infilling, zero-shot priming and fine-tuning. We propose a new framework that generalizes prior approaches, including fully unidirectional models like GPT, fully bidirectional models like BERT, and hybrid models like CM3 and prefix LM. Our framework distinguishes between two notions of bidirectionality---bidirectional context and bidirectional attention---and allows us to control each of them separately. We find that the optimal configuration is largely application-dependent (e.g., bidirectional attention is beneficial for fine-tuning and infilling, but harmful for next token prediction and zero-shot priming). We train models with up to 6.7B parameters, and find differences to remain consistent at scale. While prior work on scaling has focused on left-to-right autoregressive models, our results suggest that this approach comes with some trade-offs, and it might be worthwhile to develop very large bidirectional models.
\end{abstract}

\begin{figure*}[t]
\centering
\includegraphics[width=0.8\linewidth]{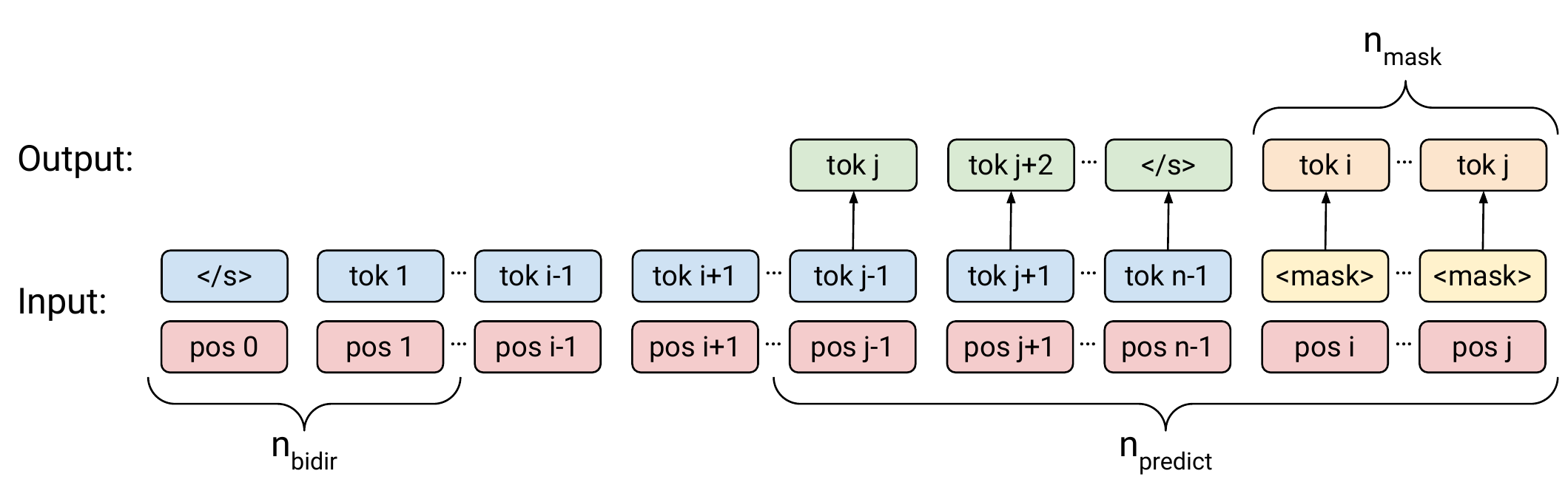}
\caption{
\textbf{Proposed framework.}
Starting from the original document, we mask $\nmask$ tokens at random and move them---along with their positional embeddings---to the end.
We define our loss over the last $\npredict$ tokens, predicting the masked token for the last $\nmask$, and the next token for the remaining $\npredict-\nmask$.
We use bidirectional attention over the first $\nbidir$ tokens, and unidirectional attention over the rest.
Refer to Appendix \ref{app:framework} for a more detailed description.
}
\label{fig:framework}
\end{figure*}

\begin{table*}[ht]
\begin{center}
\begin{small}
\resizebox{\textwidth}{!}{
\begin{tabular}{lcccl}
\toprule
Name & $\nmask$ & $\nbidir$ & $\npredict$ & Related models \\
\midrule
\clm{} & $0$ & $0$ & $n$ & GPT \citep{radford2018gpt,radford2019gpt2,brown2020gpt3} \\
\clmp{}$^\dagger$ & $0$ & $\uniform(1, n)$ & $n - \nbidir$ & Prefix LM \citep{raffel2020t5,wu2021yuan} \\
\mlmc{} & $\binomial(n, 0.15)$ & $0$ & $\nmask$ & -- \\
\mlm{} & $\binomial(n, 0.15)$ & $n$ & $\nmask$ & BERT \citep{devlin-etal-2019-bert}, RoBERTa \citep{liu2019roberta} \\
\cmlm{}$^\dagger$ & $\binomial(n, 0.15)$ & $0$ & $n$ & CM3 \citep{aghajanyan2022cm3} \\
\cmlmp{}$^\dagger$ & $\binomial(n, 0.15)$ & $\uniform(1, n)$ & $\max(n - \nbidir, \nmask)$ & -- \\
\bottomrule
\end{tabular}
}
\end{small}
\end{center}
\caption{\textbf{Variants of the proposed framework explored in this work.} $n$ denotes the document length; $\binomial(n, p)$ denotes the binomial distribution; $\uniform(a, b)$ denotes the discrete uniform distribution. $^\dagger$We set $\nbidir=0$ and $\nmask=0$ with probability $p=0.1$, so that the model gets more exposure to regular language modeling.}
\label{tab:variants}
\end{table*}

\section{Introduction}

NLP has undergone a paradigm shift driven by pre-trained models like GPT and BERT \citep{bommasani2021opportunities}. These models are trained on unlabeled corpora in a self-supervised fashion, and can be effectively adapted to downstream tasks either through conventional fine-tuning \citep{devlin-etal-2019-bert} or few-shot priming \citep{brown2020gpt3}.

Despite their widespread use, there is not a universal formula to pre-train language models: prior work has explored different architectures and learning objectives, often focusing on different applications.
For instance, BERT \citep{devlin-etal-2019-bert} pre-trained masked language models for NLU fine-tuning, BART \citep{lewis-etal-2020-bart} pre-trained seq2seq models on denoising for both NLU and generation tasks, and GPT-3 \citep{brown2020gpt3} scaled autoregressive language models focusing on zero- and few-shot priming.
However, such models differ on many factors in addition to their architecture and learning objective (e.g., the pre-training data, compute and hyperparameters), making a principled comparison difficult.
Motivated by that, \citet{raffel2020t5} presented a comprehensive study exploring various pre-training objective and architecture variants in a controlled environment. However, they conducted most of the exploration using small models, while recent work has found that different approaches behave differently at scale \citep{anonymous2021scaling,tay2022scale}, and their evaluation was limited to fine-tuning.

In this paper, we focus on a key factor that differentiates many pre-training approaches---bidirectionality---and study it in different settings as a function of scale. We propose a new framework that distinguishes between two notions of bidirectionality: \textbf{bidirectional context} (whether the prediction of a given token is conditioned on both the right and the left context, or only on either of them), and \textbf{bidirectional attention} (whether there are blocks of tokens that can all attend to each other, contrasting with triangular attention masking). Our framework offers knobs to control each of them separately, generalizing several previous approaches (e.g. BERT leverages both types of bidirectionality, GPT does not use any, prefix LMs only leverage bidirectional attention, and CM3 only leverages bidirectional context).%

We train a total of 24 models covering 6 variants of our framework and 5 model sizes with up to 6.7B parameters, and evaluate them on 4 settings: language modeling, text infilling, zero-shot priming, and fine-tuning. We find that bidirectional attention and context have a different impact depending on the use case, and there is not a single configuration that is optimal for all scenarios. Moreover, we find this behavior to remain consistent at the scale range considered in this study. With recent scaling work focusing on fully unidirectional models, this suggests that there is potential for alternative architectures and learning objectives that might be better suited for other use cases.

\section{Proposed framework}
\label{sec:framework}

As illustrated in Figure \ref{fig:framework}, we propose a generalized framework to pre-train transformer models on unlabeled corpora. Our framework supports both unidirectional and bidirectional attention, as well as next token prediction and single-token infilling, using the following \textbf{parameters} to balance them:
\begin{itemize}
    \item $\nbidir$ controls the length of the prefix using bidirectional attention, whereas the rest of the document uses unidirectional attention. More concretely, we set the attention mask so that the $i$th token can attend to the $j$th token if and only if $j \leq \max (i, \nbidir)$.
    \item $\nmask$ controls how many tokens are masked. Masked tokens are moved to the end along with their positional embeddings.
    \item $\npredict$ controls the length of the suffix for which we define our supervisory signal. We use the cross-entropy loss to train the model, predicting the masked tokens for the last $\nmask$, and the next token for the remaining $\npredict - \nmask$.\footnote{We set $\npredict \leq  n - \nbidir + \nmask$ so we only predict tokens that are either masked or cannot attend to themselves.} %
\end{itemize}
As such, our framework allows us to vary the \textbf{two notions of bidirectionality} discussed above: $\nbidir$ controls the weight of bidirectional attention, whereas $\nmask$ and $\npredict$ control the weight of bidirectional context. In addition, larger values of $\npredict$ result in more tokens of supervision.

Table \ref{tab:variants} summarizes the specific \textbf{variants} of this general framework that we explore in our experiments, along with a descriptive name that we will use to refer to each of them. Some variants are equivalent or closely related to existing approaches. In particular, \clm{} is equivalent to conventional autoregressive language models, and \clmp{} is equivalent to prefix language models. \mlm{} is closely related to the RoBERTa objective,\footnote{Moving masked tokens to the end becomes irrelevant when $\nbidir=n$, as their positional embeddings move with them and transformers operate over sets.} except that we do not replace 10\% of the masked tokens with the original or a randomly picked one. \cmlm{} is similar to the CM3 objective, except that we mask individual tokens instead of spans and we draw the number of masks from a binomial distribution. Finally, we introduce \mlmc{} as a variant of \mlm{} using unidirectional attention (or, from another perspective, a variant of \cmlm{} predicting masked tokens alone), and \cmlmp{} as a variant of \cmlm{} using a bidirectional attention prefix.

\section{Experimental settings}

\begin{table}[t]
\begin{center}
\begin{small}
\addtolength{\tabcolsep}{-2.5pt}
\begin{tabular}{rrrrrrr}
\toprule
\multicolumn{1}{c}{\emph{size}} & \multicolumn{1}{c}{\emph{cost}} & \multicolumn{1}{c}{$l$} & \multicolumn{1}{c}{$d$} & \multicolumn{1}{c}{$h$} & \multicolumn{1}{c}{\emph{bs}} & \multicolumn{1}{c}{\emph{lr}} \\
\midrule
125M & 0.11 & 12 & 768 & 12 & 0.5M & 6e-4
\\
355M & 0.31 & 24 & 1024 & 16 & 0.5M & 3e-4
\\
1.3B & 1.11 & 24 & 2048 & 32 & 1M & 2e-4
\\
2.7B & 2.23 & 32 & 2560 & 32 & 1M & 1.6e-4
\\
6.7B & 5.49 & 32 & 4096 & 32 & 2M & 1.2e-4
\\
\bottomrule
\end{tabular}%
\end{small}
\end{center}
\caption{\textbf{Model details}. \emph{size}: number of parameters, \emph{cost}: training ZFLOPs, $l$: layers, $d$: hidden dimension, $h$: attention heads, \emph{bs}: batch size, \emph{lr}: learning rate. All models are trained for 100B tokens with a maximum sequence length of 1024 tokens. We estimate training ZFLOPs analytically following \citet{artetxe2021moe}.}
\label{tab:models}
\end{table}

\subsection{Models}

For each variant in Table \ref{tab:variants}, we train models at different scales using the same settings as \citet{artetxe2021moe}, which at the same time roughly follow \citet{brown2020gpt3}. So as to reduce the computational cost of our exploration, we differ from \citet{artetxe2021moe} in two ways: (i) we use a maximum sequence length of 1024 tokens instead of 2048, and (ii) we train for 100B tokens instead of 300B. At the same time, we only train 125M and 355M models for the \clmp{} and \mlmc{} variants. Table \ref{tab:models} summarizes the settings that we use for each model.

We use the same training data as \citet{artetxe2021moe}, which combines BookCorpus \citep{zhu2015bookcorpus}, CC-News \citep{nagel2016ccnews}, OpenWebText \citep{gokaslan2019openwebtext}, CC-Stories \citep{trinh2018simple}, and English CC100 \citep{wenzek-etal-2020-ccnet}, totalling 112B tokens. %
Following them, we also use the same BPE encoding as GPT-2 \citep{radford2019gpt2} with a vocabulary of 50k.

Our implementation is based in fairseq \citep{ott-etal-2019-fairseq}. We apply the procedure described in \S\ref{sec:framework} to each document separately, and combine multiple documents into a single sequence to speed up training.\footnote{We achieve this using \texttt{---sample-break-mode complete} in fairseq. This is different from \citet{artetxe2021moe}, who concatenated all documents and split the resulting sequence into non-overlapping blocks without respecting document boundaries (\texttt{---sample-break-mode none}).} As such, we move the masked tokens to the end of each document (as opposed to the end of the whole sequence), and apply a bidirectional attention prefix to each document rather than the sequence as a whole.\footnote{As a consequence, a given token cannot attend to tokens in future documents even when $\nbidir=n$, but all tokens can attend to tokens in previous documents.}

\begin{figure*}[t]
\centering
\includegraphics[width=\linewidth]{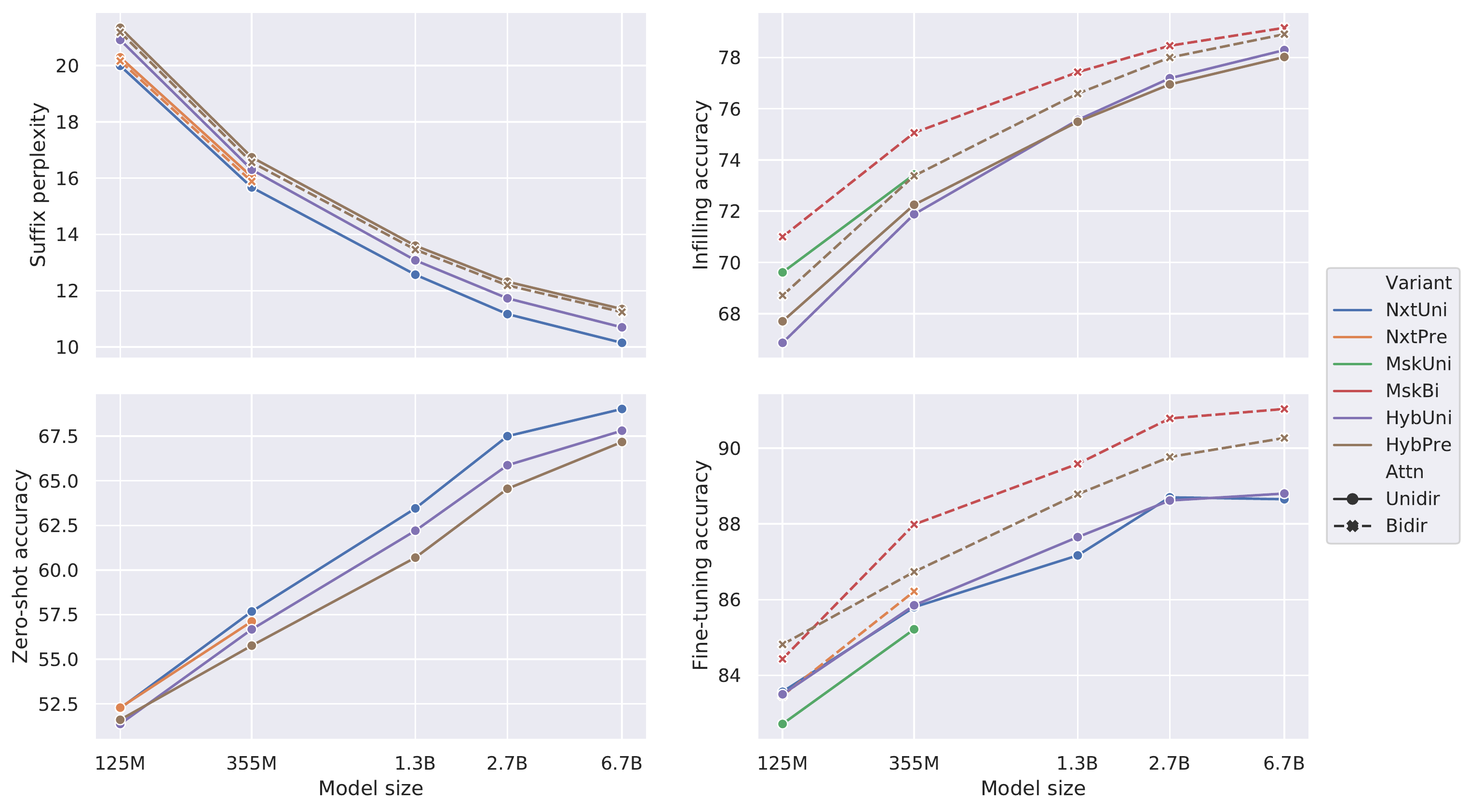}
\caption{
\textbf{Main results.} \textit{Unidir} and \textit{Bidir} denote using $\nbidir=0$ and $\nbidir=n$ after pre-training, respectively (or $\nbidir=n_\mathrm{prefix}$ for suffix perplexity).
}
\label{fig:results}
\end{figure*}

\subsection{Evaluation} \label{subsec:evaluation}

We evaluate our models in the following settings:

\paragraph{Language modeling.} We evaluate the ability of our models to predict the next token in a sequence as measured by perplexity.\footnote{We exclude \mlm{} and \mlmc{} as they are not trained on next token prediction.} Different from training, we do not concatenate different documents into the same sequence, and instead score each document as a separate sequence.\footnote{This corresponds to the \texttt{---sample-break-mode complete\_doc} option in fairseq.} Given that \clmp{} and \cmlmp{} are primarily trained to predict the last part of a document conditioned on the first part, we also measure the perplexity at predicting the last 20\% tokens in each document conditioned on the first 80\%. So as to understand whether using bidirectional attention in the prefix is useful to that end, we try different values of $\nbidir$ according to a ratio $\rbidir$, so that $\nbidir=\rbidir \times n_\mathrm{prefix}$ and $n_\mathrm{prefix}=0.8n$ is the length of the prefix we are conditioning on.

\paragraph{Single token infilling.} We mask a single word in each document at random, and measure the accuracy at predicting it.\footnote{Similar to language modeling evaluation, we feed each document as a separate sequence.} To that end, we use the same procedure used for training (illustrated in Figure \ref{fig:framework}), which moves the mask token to the end of the sequence.\footnote{For models trained with a bidirectional attention prefix, we try different values of $\rbidir$ at inference time, so that $\nbidir = \rbidir \times n$.}
This approach is not suitable for models trained exclusively on next token prediction like \clm{} and \clmp{}, as they can only be conditioned on the right context. However, one can still use such models for infilling in a generative fashion, replacing the masked token with each element in the vocabulary, scoring the resulting sequences autoregressively, and predicting the token yielding the highest scoring sequence. In addition to our primary evaluation, we compare both of these approaches, which we refer to as \textit{infill} (direct infilling) and \textit{full} (full sequence scoring). Given that \textit{full} can be prohibitively expensive when considering the full vocabulary, we constrain the set of options to the top 32 candidates generated by the 125M \mlm{} model.\footnote{The top 32 candidates contain the correct one in 95.19\% of the cases, which is the upper bound accuracy in this setting.} %

\paragraph{Zero-shot priming.} We evaluate our models on zero-shot priming using the exact same settings and tasks as \citet{artetxe2021moe}, which comprises ReCoRD~\citep{zhang2018record}, HellaSwag~\citep{zellers-etal-2019-hellaswag}, PIQA~\citep{bisk2020piqa}, WinoGrande~\citep{sakaguchi2020winogrande}, StoryCloze~\citep{mostafazadeh-etal-2016-corpus} and OpenBookQA~\citep{mihaylov-etal-2018-suit}.
These are all multiple choice tasks, so we score the populated prompt corresponding to each option in an autoregressive fashion and predict the highest scoring one.\footnote{Refer to \citet{artetxe2021moe} for a description of the scoring function used for each task and the evaluatio protocol.} However, when the options differ in a single token---as it is common for classification tasks with single-token verbalizers---one can also score such token directly in an infilling fashion. So as to understand how both approaches compare, we further evaluate our models on MNLI \citep{williams-etal-2018-broad}, using a single-token verbalizer placed in the middle of the prompt.\footnote{We use \texttt{<premise>, right? \{Yes|No|Also\}, <hypothesis>} as our template and report results on the matched development set.}

\paragraph{Fine-tuning.} We experiment with the following tasks from GLUE \citep{wang2018glue}: COLA \citep{warstadt-etal-2019-neural}, MNLI-m \citep{williams-etal-2018-broad}, MRPC \citep{dolan-brockett-2005-automatically}, QNLI \citep{rajpurkar-etal-2016-squad}, RTE \citep{dagan2006rte1,haim2006rte2,giampiccolo-etal-2007-third,bentivogli2009rte5} and SST-2 \citep{socher-etal-2013-recursive}. Our fine-tuning approach closely follows BERT and similar models: we place a special \texttt{</s>} token at the end of the sequence (analogous to the special \texttt{<CLS>} token used by BERT) and learn a new classification head on top. We ran a grid search with the learning rate in \{1e-0.5, 2e-05, 5e-05, 5e-06\} and batch size in \{16, 32, 64\}, and report the best development accuracy for each model. The rest of hyperparameters follow RoBERTa. For all variants, we tried fine-tuning both with fully unidirectional attention ($\rbidir=0)$ and fully bidirectional attention ($\rbidir=1$). Refer to Appendix \ref{app:finetuning} for more details.

\section{Results}

We visualize our main results in Figure \ref{fig:results}, and discuss each setting in more detail next.

\begin{table}[t]
\begin{center}
\begin{small}
\begin{tabular}{lccccc}
\toprule
 & 125M & 355M & 1.3B & 2.7B & 6.7B \\
\midrule
\clm{} & \textbf{22.23} & \textbf{17.49} & \textbf{14.07} & \textbf{12.55} & \textbf{11.44} \\
\clmp{} & 22.75 & 18.06 & -- & -- & -- \\
\cmlm{} & 23.26 & 18.19 & 14.65 & 13.16 & 12.03 \\
\cmlmp{} & 23.91 & 18.81 & 15.33 & 13.92 & 12.86 \\
\bottomrule
\end{tabular}%
\end{small}
\end{center}
\caption{\textbf{Full document perplexity}.}
\label{tab:clm}
\end{table}

\begin{table}[t]
\begin{center}
\begin{small}
\addtolength{\tabcolsep}{-2.5pt}
\begin{tabular}{lcccccc}
\toprule
& $\rbidir$ & 125M & 355M & 1.3B & 2.7B & 6.7B \\
\midrule
\clm{} & 0.00 & \textbf{19.99} & \textbf{15.67} & \textbf{12.57} & \textbf{11.17} & \textbf{10.15} \\
\midrule
\multirow{5}{*}{\clmp{}}
& 0.00 & 20.29 & 16.05 & -- & -- & -- \\
& 0.25 & 20.25 & 16.00 & -- & -- & -- \\
& 0.50 & 20.21 & 15.96 & -- & -- & -- \\
& 0.75 & 20.17 & 15.92 & -- & -- & -- \\
& 1.00 & \underline{20.16} & \underline{15.88} & -- & -- & -- \\
\midrule
\cmlm{} & 0.00 & 20.91 & 16.30 & 13.08 & 11.73 & 10.70 \\
\midrule
\multirow{5}{*}{\cmlmp{}}
& 0.00 & 21.34 & 16.74 & 13.60 & 12.32 & 11.35 \\
& 0.25 & 21.30 & 16.69 & 13.56 & 12.29 & 11.33 \\
& 0.50 & 21.26 & 16.66 & 13.54 & 12.26 & 11.30 \\
& 0.75 & 21.23 & 16.62 & 13.51 & 12.23 & 11.28 \\
& 1.00 & \underline{21.18} & \underline{16.56} & \underline{13.46} & \underline{12.19} & \underline{11.24} \\
\bottomrule
\end{tabular}%
\end{small}
\end{center}
\caption{\textbf{Suffix perplexity}. We measure perplexity at predicting the last 20\% of the tokens in each document conditioned on the first 80\%, using $\nbidir=\rbidir \times n_\mathrm{prefix}$ for inference, where $n_\mathrm{prefix}=0.8n$ is the length of the prefix we are conditioning on.}
\label{tab:suffix}
\end{table}

\subsection{Language modeling} \label{subsec:clm}

We report full document perplexities in Table \ref{tab:clm}. \clm{} obtains the best results, followed by \cmlm{} and \cmlmp{}, and \clmp{} doing slightly better than \cmlm{} at small scale. This is consistent with how close the pre-training objective is to the end task: \clm{} is exclusively trained on next token prediction, \cmlm{} combines it with masking (which is not used here), and \cmlmp{} further combines it with a bidirectional attention prefix (which is not used here either). However, it is interesting that scaling up does not reduce the gap between them. This suggests that there is some fundamental interference between these different capabilities,\footnote{There are various factors that could explain this. Both masking and the bidirectional attention prefix reduce the supervision on next token prediction, and masking further introduces some noise in the original sequence. Moreover, training to use both unidirectional and bidirectional attention and/or context might provide a conflicting signal, although our results later in \S\ref{subsec:infilling} suggest that this does not have a major impact at scale.} and increasing capacity does not mitigate it.  %

Table \ref{tab:suffix} reports suffix perplexity results, where we predict the last 20\% of the tokens in each document conditioned on the rest. Compared to the previous results, \clmp{} and \cmlmp{} reduce the gap with \clm{} and \cmlm{}, but they still lag behind them. In both cases, we find that the models benefit from using bidirectional attention in the prefix at inference time (i.e., higher values of $\rbidir$ yield lower perplexity), but the improvement is relatively small. It is intriguing that \clm{} outperforms \clmp{}, when the latter was trained on suffix prediction and can leverage bidirectional attention. We attribute this to the bidirectional prefix reducing the number of tokens of supervision during training. %

\begin{table}[t]
\begin{center}
\begin{small}
\addtolength{\tabcolsep}{-2.5pt}
\begin{tabular}{lcccccc}
\toprule
& $\rbidir$ & 125M & 355M & 1.3B & 2.7B & 6.7B \\
\midrule
\mlmc{} & 0.00 & 69.61 & 73.43 & -- & -- & -- \\
\midrule
\mlm{} & 1.00 & \textbf{71.00} & \textbf{75.06} & \textbf{77.43} & \textbf{78.46} & \textbf{79.16} \\
\midrule
\cmlm{} & 0.00 & 66.86 & 71.88 & 75.56 & 77.19 & 78.29 \\
\midrule
\multirow{5}{*}{\cmlmp{}}
& 0.00 & 67.70 & 72.25 & 75.49 & 76.95 & 78.02 \\
& 0.25 & 68.02 & 72.57 & 75.77 & 77.25 & 78.22 \\
& 0.50 & 68.23 & 72.85 & 76.05 & 77.48 & 78.52 \\
& 0.75 & 68.47 & 73.13 & 76.32 & 77.74 & 78.70 \\
& 1.00 & \underline{68.71} & \underline{73.38} & \underline{76.59} & \underline{78.00} & \underline{78.91} \\
\bottomrule
\end{tabular}%
\end{small}
\end{center}
\caption{\textbf{Single token infilling accuracy}. We mask a random token in each validation document and measure the accuracy at predicting it, using $\nbidir=\rbidir \times n$ for inference.}
\label{tab:infilling}
\end{table}

\subsection{Single token infilling} \label{subsec:infilling}

We report infilling results in Table \ref{tab:infilling}. \mlm{} obtains the best results, which can be explained by its use of bidirectional attention and the fact that it is exclusively trained on masking. Our results suggest that both of these factors play a role, but their impact varies at scale. As for the first factor, we find that bidirectional attention has a larger impact on infilling compared to next token prediction (\S\ref{subsec:clm}), as reflected by \mlm{} doing substantially better than \mlmc{}. %
Moreover, we find that this also holds at scale, as reflected by \cmlmp{} doing better with larger values of $\rbidir$, while outperforming \cmlm{}. Regarding the second factor, we find that combining masking with next token prediction significantly hurts infilling performance for small models, as reflected by the large gap between \mlmc{} and \cmlm{}. %
However, we also find the impact of this to vanish at scale, as reflected by the gap between \mlm{} and \cmlmp{} with $\rbidir=1.0$ becoming smaller for larger models. This also explains why \cmlmp{} with $\rbidir=0.0$ outperforms \cmlm{} for small models, but the trend is reversed as we scale up: the bidirectional prefix in \cmlmp{} reduces the relative weight of next token prediction during training, which outweighs the discrepancy with not using bidirectional attention at inference time for small models, but not for larger ones.
Interestingly, this is different from the behavior observed for language modeling in \S\ref{subsec:clm}, where scale did not significantly mitigate the negative impact of combining masking and next token prediction during training. We attribute this to masking introducing noise in the original document, as well as reducing the amount of tokens that we train on next token prediction.\footnote{Note that the reverse is not true: the addition of next token prediction in \cmlm{} does not reduce the amount of supervision on infilling with respect to \mlmc{}, as we use the same value of $\nmask$ in both cases.}

\begin{table}[t]
\begin{center}
\begin{small}
\addtolength{\tabcolsep}{-2.5pt}
\begin{tabular}{llccccc}
\toprule
&  & 125M & 355M & 1.3B & 2.7B & 6.7B \\
\midrule
\clm{} & full & 69.83 & 73.13 & 75.90 & 77.26 & 77.98 \\
\midrule
\clmp{} & full & 69.40 & 72.75 & -- & -- & -- \\
\midrule
\mlmc{} & infill & 69.65 & 73.39 & -- & -- & -- \\
\midrule
\mlm{} & infill$^\dagger$ & \textbf{71.00} & \textbf{74.98} & \textbf{77.17} & \textbf{78.07} & \textbf{78.70} \\
\midrule
\multirow{2}{*}{\cmlm{}}
& full & \underline{68.94} & \underline{72.77} & \underline{75.43} & 76.61 & 77.76 \\
& infill & 67.02 & 71.90 & 75.38 & \underline{76.90} & \underline{77.88} \\
\midrule
\multirow{3}{*}{\cmlmp{}}
& full & 68.53 & 72.05 & 74.75 & 76.03 & 76.87 \\
& infill & 67.82 & 72.24 & 75.35 & 76.66 & 77.63 \\
& infill$^\dagger$ & \underline{68.78} & \underline{73.35} & \underline{76.36} & \underline{77.63} & \underline{78.47} \\
\bottomrule
\end{tabular}%
\end{small}
\end{center}
\caption{\textbf{Single token infilling accuracy, re-ranking the top 32 candidates from 125M \mlm{}}. $^\dagger$ denotes $\nbidir=n$, the rest use $\nbidir=0$. Refer to \S\ref{subsec:evaluation} for more details.}
\label{tab:infilling_top32}
\end{table}

Table \ref{tab:infilling_top32} reports infilling results re-ranking the top 32 candidates from the 125M \mlm{} model. The best results are still obtained by \mlm{}, but we find the generative approach described in \S\ref{subsec:evaluation} to be competitive, with \clm{} obtaining the second best results at 125M and the third best results for larger models. This suggests that models trained exclusively on next token prediction can also be used for infilling as long as the set of candidates is small, even outperforming hybrid models like \cmlm{} that are trained both on next token prediction and infilling itself. In fact, it is remarkable that \clm{} is only outperformed by models using bidirectional attention which, consistent with our previous results, seems strongly beneficial for infilling. Nevertheless, we also find direct infilling (\textit{infill}) to scale better than generative full sequence scoring (\textit{full}) for both \cmlm{} and \cmlmp{}, although this could (partly) be explained by the interference between next token prediction and masking diminishing at scale as discussed previously.

\begin{table}[t]
\begin{center}
\begin{small}
\addtolength{\tabcolsep}{-2.5pt}
\resizebox{0.48\textwidth}{!}{
\begin{tabular}{clcccccc|c}
\toprule
&& RE & HS & PI & WG & SC & OB & avg \\
\midrule
\multirow{4}{*}{125M}
& \clm{} & \underline{66.7} & \underline{32.2} & \underline{65.3} & 51.9 & \underline{64.3} & 33.0 & \underline{52.3} \\
& \clmp{} & 65.8 & 31.2 & 64.1 & \underline{54.1} & 63.5 & 35.0 & \underline{52.3} \\
& \cmlm{} & 65.4 & 30.8 & 63.1 & 50.9 & 63.6 & 34.4 & 51.4 \\
& \cmlmp{} & 64.9 & 30.5 & 64.2 & 51.9 & 63.0 & \underline{35.2} & 51.6 \\
\midrule
\multirow{4}{*}{355M}
& \clm{} & \underline{74.8} & \underline{41.0} & \underline{69.5} & 52.2 & \underline{70.0} & \underline{38.6} & \underline{57.7} \\
& \clmp{} & 74.3 & 40.0 & 68.9 & \underline{52.6} & 69.2 & 37.8 & 57.1 \\
& \cmlm{} & 73.9 & 39.3 & 68.1 & 52.3 & 69.3 & 37.2 & 56.7 \\
& \cmlmp{} & 72.9 & 37.8 & 67.6 & 50.4 & 68.4 & 37.4 & 55.8 \\
\midrule
\multirow{3}{*}{1.3B}
& \clm{} & \underline{81.0} & \underline{52.6} & \underline{73.8} & \underline{55.6} & \underline{74.1} & \underline{43.6} & \underline{63.5} \\
& \cmlm{} & 80.0 & 50.3 & 72.1 & 53.7 & \underline{74.1} & 43.0 & 62.2 \\
& \cmlmp{} & 79.4 & 48.5 & 71.4 & 52.9 & 73.9 & 38.2 & 60.7 \\
\midrule
\multirow{3}{*}{2.7B}
& \clm{} & \underline{83.8} & \underline{58.8} & \underline{75.0} & \underline{\textbf{60.1}} & 76.6 & \underline{50.8} & \underline{67.5} \\
& \cmlm{} & 83.1 & 57.5 & 73.9 & 58.0 & \underline{76.9} & 45.8 & 65.9 \\
& \cmlmp{} & 81.7 & 54.7 & 72.4 & 56.7 & 75.3 & 46.6 & 64.6 \\
\midrule
\multirow{3}{*}{6.7B}
& \clm{} & \underline{\textbf{85.2}} & \underline{\textbf{63.6}} & \underline{\textbf{76.2}} & \underline{60.0} & \underline{\textbf{77.6}} & \underline{\textbf{51.6}} & \underline{\textbf{69.0}} \\
& \cmlm{} & 84.2 & 61.7 & 75.5 & 59.7 & 76.8 & 49.0 & 67.8 \\
& \cmlmp{} & 83.9 & 58.9 & 73.9 & 58.7 & 76.9 & 50.8 & 67.2 \\
\bottomrule
\end{tabular}}
\end{small}
\end{center}
\caption{\textbf{Zero-shot priming accuracy}. We use $\nbidir=0$ for inference. \texttt{RE}: ReCoRD, \texttt{HS}: HellaSwag, \texttt{PI}: PIQA, \texttt{WG}: WinoGrande, \texttt{SC}: StoryCloze, \texttt{OB}: OpenBookQA.}
\label{tab:zeroshot}
\end{table}

\begin{table}[t]
\begin{center}
\begin{small}
\addtolength{\tabcolsep}{-2.5pt}
\begin{tabular}{llccccc}
\toprule
&  & 125M & 355M & 1.3B & 2.7B & 6.7B \\
\midrule
\clm{} & full & 44.79 & \textbf{50.12} & \textbf{53.63} & 55.09 & \textbf{55.27} \\
\midrule
\clmp{} & full & \textbf{45.41} & 49.15 & -- & -- & -- \\
\midrule
\mlmc{} & infill & 41.69 & 44.15 & -- & -- & -- \\
\midrule
\mlm{} & infill$^\dagger$ & 41.56 & 48.34 & 52.24 & \textbf{55.59} & 53.97 \\
\midrule
\multirow{2}{*}{\cmlm{}}
& full & \underline{45.12} & \underline{47.92} & \underline{52.59} & \underline{53.40} & \underline{54.47} \\
& infill & 43.03 & 44.54 & 48.13 & 49.94 & 51.26 \\
\midrule
\multirow{3}{*}{\cmlmp{}}
& full & \underline{43.37} & \underline{47.54} & \underline{51.53} & \underline{52.36} & \underline{54.01} \\
& infill & 42.16 & 44.47 & 47.36 & 49.98 & 50.24 \\
& infill$^\dagger$ & 42.95 & 46.57 & 49.13 & 51.85 & 52.41 \\
\bottomrule
\end{tabular}%
\end{small}
\end{center}
\caption{\textbf{Zero-shot MNLI accuracy}. $^\dagger$ denotes $\nbidir=n$, the rest use $\nbidir=0$.}
\label{tab:zeroshot_mnli}
\end{table}

\subsection{Zero-shot priming}

We report zero-shot priming results in Table \ref{tab:zeroshot}. We observe the same general trends as in language modeling (\S\ref{subsec:clm}), with \clm{} performing best, followed by \cmlm{} and \cmlmp{}. The results are generally consistent across tasks.%

Table \ref{tab:zeroshot_mnli} reports MNLI results, comparing full sequence scoring and direct infilling. Consistent with the intrinsic evaluation in \S\ref{subsec:infilling}, we find full sequence scoring with \clm{} to be competitive with direct infilling with \mlm{}. In fact, full sequence scoring does even better comparatively, obtaining the best results in all but one of the model sizes. Moreover, it is remarkable that both \cmlm{} and \cmlmp{} obtain better results with full sequence scoring compared to direct infilling in all cases. Consistent with our previous results, this suggests that left-to-right language models can be a valid or even superior alternative to masked language models for single-token infilling tasks, as long as one can afford scoring each candidate separately.

\begin{table}[t]
\begin{center}
\begin{small}
\addtolength{\tabcolsep}{-2.5pt}
\begin{tabular}{lcccccc}
\toprule
& $\rbidir$ & 125M & 355M & 1.3B & 2.7B & 6.7B \\
\midrule
\multirow{2}{*}{\clm{}}
& 0.0 & \underline{83.6} & \underline{85.8} & \underline{87.2} & \underline{88.7} & \underline{88.6} \\
& 1.0 & 75.9 & 77.1 & 79.0 & 79.2 & 80.3 \\
\midrule
\multirow{2}{*}{\clmp{}}
& 0.0 & \underline{84.2} & 85.8 & -- & -- & -- \\
& 1.0 & 83.5 & \underline{86.2} & -- & -- & -- \\
\midrule
\multirow{2}{*}{\mlmc{}}
& 0.0 & 82.7 & \underline{85.2} & -- & -- & -- \\
& 1.0 & \underline{83.2} & 85.1 & -- & -- & -- \\
\midrule
\multirow{2}{*}{\mlm{}}
& 0.0 & 79.6 & 81.0 & 81.9 & 81.6 & 82.6 \\
& 1.0 & \underline{84.4} & \underline{\textbf{88.0}} & \underline{\textbf{89.6}} & \underline{\textbf{90.8}} & \underline{\textbf{91.0}} \\
\midrule
\multirow{2}{*}{\cmlm{}}
& 0.0 & \underline{83.5} & \underline{85.9} & \underline{87.6} & \underline{88.6} & \underline{88.8} \\
& 1.0 & 80.8 & 82.5 & 84.0 & 85.0 & 84.7 \\
\midrule
\multirow{2}{*}{\cmlmp{}}
& 0.0 & 83.6 & 86.1 & 87.1 & 88.2 & 88.2 \\
& 1.0 & \underline{\textbf{84.8}} & \underline{86.7} & \underline{88.8} & \underline{89.8} & \underline{90.3} \\
\bottomrule
\end{tabular}%
\end{small}
\end{center}
\caption{\textbf{Average fine-tuning accuracy}.}
\label{tab:finetuning_bidir}
\end{table}

\begin{table*}[ht]
\begin{center}
\begin{small}
\addtolength{\tabcolsep}{-2.5pt}
\begin{tabular}{clcccccc|c}
\toprule
&& COLA & MNLI & MRPC & QNLI & RTE & SST2 & avg \\
\midrule

\multirow{6}{*}{125M}
& \clm{} & 82.4 & 83.1 & 82.8 & 88.8 & 70.4 & \underline{93.9} & 83.6 \\
& \clmp{} & 81.3 & 83.3 & 83.1 & 90.1 & 69.3 & 93.7 & 83.5 \\
& \mlmc{} & 82.6 & 82.2 & 81.4 & 88.4 & 68.6 & 93.1 & 82.7 \\
& \mlm{} & \underline{83.2} & \underline{84.8} & \underline{85.5} & \underline{91.0} & 68.6 & 93.5 & 84.4 \\
& \cmlm{} & 82.7 & 83.1 & 83.6 & 89.3 & 69.3 & 93.0 & 83.5 \\
& \cmlmp{} & 82.5 & 84.2 & \underline{85.5} & 90.9 & \underline{72.6} & 93.2 & \underline{84.8} \\
\midrule
\multirow{6}{*}{355M}
& \clm{} & 84.2 & 85.8 & 84.1 & 91.2 & 74.7 & 94.8 & 85.8 \\
& \clmp{} & 83.8 & 86.3 & 86.5 & 92.0 & 73.3 & 95.4 & 86.2 \\
& \mlmc{} & 84.0 & 84.4 & 84.6 & 90.5 & 73.6 & 94.2 & 85.2 \\
& \mlm{} & 85.2 & \underline{87.7} & \underline{89.7} & \underline{92.9} & \underline{76.2} & \underline{96.2} & \underline{88.0} \\
& \cmlm{} & \underline{85.4} & 85.3 & 85.3 & 91.0 & 73.3 & 94.8 & 85.9 \\
& \cmlmp{} & 84.5 & 86.5 & 87.3 & 92.5 & 74.4 & 95.2 & 86.7 \\
\midrule
\multirow{4}{*}{1.3B}
& \clm{} & \underline{87.0} & 87.3 & 85.3 & 92.4 & 75.1 & 95.9 & 87.2 \\
& \mlm{} & 85.7 & \underline{89.1} & 89.7 & \underline{93.9} & \underline{82.3} & \underline{\textbf{96.8}} & \underline{89.6} \\
& \cmlm{} & 86.3 & 87.0 & 86.0 & 92.3 & 78.0 & 96.3 & 87.6 \\
& \cmlmp{} & 85.1 & 88.4 & \underline{90.0} & 93.6 & 79.4 & 96.2 & 88.8 \\
\midrule
\multirow{4}{*}{2.7B}
& \clm{} & 86.0 & 88.5 & 85.5 & 93.0 & 83.0 & 96.2 & 88.7 \\
& \mlm{} & \underline{\textbf{87.2}} & \underline{\textbf{89.8}} & \underline{\textbf{91.7}} & 94.0 & \underline{85.2} & \underline{\textbf{96.8}} & \underline{90.8} \\
& \cmlm{} & 86.2 & 88.1 & 86.8 & 93.0 & 80.9 & 96.7 & 88.6 \\
& \cmlmp{} & 86.2 & 89.4 & 89.5 & \underline{94.1} & 82.7 & 96.7 & 89.8 \\
\midrule
\multirow{4}{*}{6.7B}
& \clm{} & 86.3 & 88.5 & 85.8 & 93.4 & 81.2 & 96.7 & 88.6 \\
& \mlm{} & \underline{86.7} & \underline{89.6} & \underline{90.9} & \underline{\textbf{94.5}} & \underline{\textbf{87.7}} & \underline{\textbf{96.8}} & \underline{\textbf{91.0}} \\
& \cmlm{} & \underline{86.7} & 88.4 & 87.7 & 93.4 & 80.5 & 96.1 & 88.8 \\
& \cmlmp{} & 86.0 & 89.5 & 89.5 & 94.3 & 85.6 & 96.7 & 90.3 \\
\bottomrule
\end{tabular}%
\end{small}
\end{center}
\caption{\textbf{Fine-tuning accuracy}. We use $\nbidir=0$ for \clm{}, \mlmc{} and \cmlm{}, and $\nbidir=n$ for the rest.
}
\label{tab:finetuning}
\end{table*}

\subsection{Fine-tuning}

We report average fine-tuning results comparing unidirectional and bidirectional attention in Table \ref{tab:finetuning_bidir}, and full results for the optimal setting for each variant in Table \ref{tab:finetuning}.

Our results show that bidirectional attention is helpful for fine-tuning regardless of scale, with fully bidirectional models (\mlm{}) performing the best, followed by models pre-trained with a bidirectional attention prefix (\cmlmp{}, \clmp{}), and fully unidirectional models performing the worst (\cmlm{}, \clm{}, \mlmc{}). Interestingly, changing the attention type at fine-tuning time (using unidirectional attention for pre-training and bidirectional attention for fine-tuning, or the other way around) works poorly.

At the same time, we find that the role of bidirectional context is dependant on the type of attention used. When using fully unidirectional attention, bidirectional context has no clear impact, with \clm{} and \cmlm{} performing similarly. In contrast, when using bidirectional attention, bidirectional context seems beneficial, with \cmlmp{} performing better than \clmp{} at small scale. This suggests that pre-training with bidirectional context is important for the model to learn to make effective use of bidirectional attention.%

\section{Related work}

While it was once common to use random initialization for supervised learning, a series of works showed substantial improvements from pre-training autoregressive models on next token prediction \citep{dai2015semisupervised,peters-etal-2018-deep,howard-ruder-2018-universal,radford2018gpt}. The \textit{pre-train/fine-tune} paradigm was further popularized by BERT \citep{devlin-etal-2019-bert} and its derivatives like RoBERTa \citep{liu2019roberta}, which obtained further gains from pre-training bidirectional encoders on masked language modeling. Subsequent work explored masking spans instead of individual tokens, using either bidirectional encoder-only models \citep{joshi-etal-2020-spanbert} or encoder-decoder models \citep{lewis-etal-2020-bart,raffel2020t5}. More recently, there has been a reborn interest on scaling left-to-right autoregressive language models with a focus on few-shot priming \citep{radford2019gpt2,brown2020gpt3,rae2021gopher,hoffmann2022chinchilla,smith2022mtnlg,chowdhery2022palm,zhang2022opt}.

While unidirectional and bidirectional models have largely been developed as separate strains of work serving a different purpose, there have also been some attempts to combine the best of both worlds. XLNet \citep{yang2019xlnet} pre-trained autoregressive models over all permutations of the factorization order, enabling the model to use bidirectional context with strong results on fine-tuning. Similarly, CM3 \citep{aghajanyan2022cm3} trained left-to-right autoregressive models, masking some spans that are predicted at the end of the sequence. ERNIE 3.0 \citep{sun2021ernie3} proposed a modular architecture, combining a shared unidirectional module with either another unidirectional module for NLG or a bidirectional module for NLU. Finally, \citet{raffel2020t5} and \citet{wu2021yuan} explored splitting documents in two halves and predicting the second one conditioned on the first one, using unidirectional attention for the former and bidirectional attention for the latter.

Despite the large body of work on language model pre-training, there is little work comparing different approaches in a systematic manner. As a notable exception, \citet{raffel2020t5} compared various architectures and learning objectives with a focus on fine-tuning. Concurrent to our work, \citet{wang2022language} conduct a comprehensive study with a focus on zero-shot learning and multi-task fine-tuning. In contrast, we focus on the specific role of bidirectionality, and compare models of different sizes.

\section{Conclusions}

In this work, we study the role of bidirectionality in language model pre-training through a new framework that generalizes previous approaches.
Our main findings are as follows:
\begin{itemize}
\item \textbf{Bidirectional attention} is strongly beneficial for infilling and fine-tuning. In contrast, prefix language models lag behind regular language models on next token prediction, even if they get a small benefit from leveraging bidirectional attention in the prefix. This behavior is consistent at scale.
\item Models trained jointly to use unidirectional and \textbf{bidirectional context}, like \cmlm{}, lag behind regular language models on next token prediction, and scale does not mitigate this. Such models also lag behind pure masked language models on infilling, but scale does help close this gap as long as they are trained with a bidirectional attention prefix. For fine-tuning, bidirectional context is beneficial when used in conjunction with bidirectional attention, but not when used with unidirectional attention.
\item While direct \textbf{infilling} requires bidirectional context and benefits from bidirectional attention as discussed above, models using unidirectional context and attention are also competitive in infilling when one can separately score each candidate. For settings where the set of candidates is small (e.g., zero-shot priming for classification), regular language models obtain comparable or even superior results to models pre-trained on infilling.
\end{itemize}

All in all, our results show that there is not a single configuration that is optimal for all use cases, and this remains generally consistent within the scale range explored in this work. While prior work on scaling has focused on left-to-right autoregressive models, this suggests that there might be other objectives and architectures that are better suited for other applications like fine-tuning. Given the cost of pre-training several models, we would like to explore modular \citep{sun2021ernie3} or adaptation \citep{wang2022language} approaches in the future, where one would either have a single model with modular components specialized for different use cases, or efficiently adapt an existing model by changing the parameters in our framework instead of training several models from scratch.

\section*{Limitations}

Our study focuses on the role of bidirectionality on language model pre-training, and does not explore other factors that might affect model performance. In particular, we mask individual tokens without considering longer spans, and do not explore the impact of the masking rate. In addition, we do not consider sequence-to-sequence models in our study, which combine bidirectional attention in the encoder and unidirectional attention in the decoder. Finally, we train all variants for the same number of tokens, making them comparable in terms of training cost, but resulting in models using a bidirectional attention prefix or a masking objective seeing less tokens of supervision.

\bibliography{anthology,custom}

\begin{thebibliography}{47}
\expandafter\ifx\csname natexlab\endcsname\relax\def\natexlab#1{#1}\fi

\bibitem[{Aghajanyan et~al.(2022)Aghajanyan, Huang, Ross, Karpukhin, Xu, Goyal,
  Okhonko, Joshi, Ghosh, Lewis, and Zettlemoyer}]{aghajanyan2022cm3}
Armen Aghajanyan, Bernie Huang, Candace Ross, Vladimir Karpukhin, Hu~Xu, Naman
  Goyal, Dmytro Okhonko, Mandar Joshi, Gargi Ghosh, Mike Lewis, and Luke
  Zettlemoyer. 2022.
\newblock \href {https://doi.org/10.48550/ARXIV.2201.07520} {Cm3: A causal
  masked multimodal model of the internet}.

\bibitem[{Artetxe et~al.(2021)Artetxe, Bhosale, Goyal, Mihaylov, Ott, Shleifer,
  Lin, Du, Iyer, Pasunuru, Anantharaman, Li, Chen, Akin, Baines, Martin, Zhou,
  Koura, O'Horo, Wang, Zettlemoyer, Diab, Kozareva, and
  Stoyanov}]{artetxe2021moe}
Mikel Artetxe, Shruti Bhosale, Naman Goyal, Todor Mihaylov, Myle Ott, Sam
  Shleifer, Xi~Victoria Lin, Jingfei Du, Srinivasan Iyer, Ramakanth Pasunuru,
  Giri Anantharaman, Xian Li, Shuohui Chen, Halil Akin, Mandeep Baines, Louis
  Martin, Xing Zhou, Punit~Singh Koura, Brian O'Horo, Jeff Wang, Luke
  Zettlemoyer, Mona Diab, Zornitsa Kozareva, and Ves Stoyanov. 2021.
\newblock \href {https://doi.org/10.48550/ARXIV.2112.10684} {Efficient large
  scale language modeling with mixtures of experts}.

\bibitem[{Bentivogli et~al.(2009)Bentivogli, Clark, Dagan, and
  Giampiccolo}]{bentivogli2009rte5}
Luisa Bentivogli, Peter Clark, Ido Dagan, and Danilo Giampiccolo. 2009.
\newblock The fifth pascal recognizing textual entailment challenge.
\newblock In \emph{TAC}.

\bibitem[{Bisk et~al.(2020)Bisk, Zellers, Le~bras, Gao, and
  Choi}]{bisk2020piqa}
Yonatan Bisk, Rowan Zellers, Ronan Le~bras, Jianfeng Gao, and Yejin Choi. 2020.
\newblock \href {https://doi.org/10.1609/aaai.v34i05.6239} {Piqa: Reasoning
  about physical commonsense in natural language}.
\newblock \emph{Proceedings of the AAAI Conference on Artificial Intelligence},
  34(05):7432--7439.

\bibitem[{Bommasani et~al.(2021)Bommasani, Hudson, Adeli, Altman, Arora, von
  Arx, Bernstein, Bohg, Bosselut, Brunskill, Brynjolfsson, Buch, Card,
  Castellon, Chatterji, Chen, Creel, Davis, Demszky, Donahue, Doumbouya,
  Durmus, Ermon, Etchemendy, Ethayarajh, Fei-Fei, Finn, Gale, Gillespie, Goel,
  Goodman, Grossman, Guha, Hashimoto, Henderson, Hewitt, Ho, Hong, Hsu, Huang,
  Icard, Jain, Jurafsky, Kalluri, Karamcheti, Keeling, Khani, Khattab, Koh,
  Krass, Krishna, Kuditipudi, Kumar, Ladhak, Lee, Lee, Leskovec, Levent, Li,
  Li, Ma, Malik, Manning, Mirchandani, Mitchell, Munyikwa, Nair, Narayan,
  Narayanan, Newman, Nie, Niebles, Nilforoshan, Nyarko, Ogut, Orr,
  Papadimitriou, Park, Piech, Portelance, Potts, Raghunathan, Reich, Ren, Rong,
  Roohani, Ruiz, Ryan, Ré, Sadigh, Sagawa, Santhanam, Shih, Srinivasan,
  Tamkin, Taori, Thomas, Tramèr, Wang, Wang, Wu, Wu, Wu, Xie, Yasunaga, You,
  Zaharia, Zhang, Zhang, Zhang, Zhang, Zheng, Zhou, and
  Liang}]{bommasani2021opportunities}
Rishi Bommasani, Drew~A. Hudson, Ehsan Adeli, Russ Altman, Simran Arora, Sydney
  von Arx, Michael~S. Bernstein, Jeannette Bohg, Antoine Bosselut, Emma
  Brunskill, Erik Brynjolfsson, Shyamal Buch, Dallas Card, Rodrigo Castellon,
  Niladri Chatterji, Annie Chen, Kathleen Creel, Jared~Quincy Davis, Dora
  Demszky, Chris Donahue, Moussa Doumbouya, Esin Durmus, Stefano Ermon, John
  Etchemendy, Kawin Ethayarajh, Li~Fei-Fei, Chelsea Finn, Trevor Gale, Lauren
  Gillespie, Karan Goel, Noah Goodman, Shelby Grossman, Neel Guha, Tatsunori
  Hashimoto, Peter Henderson, John Hewitt, Daniel~E. Ho, Jenny Hong, Kyle Hsu,
  Jing Huang, Thomas Icard, Saahil Jain, Dan Jurafsky, Pratyusha Kalluri,
  Siddharth Karamcheti, Geoff Keeling, Fereshte Khani, Omar Khattab, Pang~Wei
  Koh, Mark Krass, Ranjay Krishna, Rohith Kuditipudi, Ananya Kumar, Faisal
  Ladhak, Mina Lee, Tony Lee, Jure Leskovec, Isabelle Levent, Xiang~Lisa Li,
  Xuechen Li, Tengyu Ma, Ali Malik, Christopher~D. Manning, Suvir Mirchandani,
  Eric Mitchell, Zanele Munyikwa, Suraj Nair, Avanika Narayan, Deepak
  Narayanan, Ben Newman, Allen Nie, Juan~Carlos Niebles, Hamed Nilforoshan,
  Julian Nyarko, Giray Ogut, Laurel Orr, Isabel Papadimitriou, Joon~Sung Park,
  Chris Piech, Eva Portelance, Christopher Potts, Aditi Raghunathan, Rob Reich,
  Hongyu Ren, Frieda Rong, Yusuf Roohani, Camilo Ruiz, Jack Ryan, Christopher
  Ré, Dorsa Sadigh, Shiori Sagawa, Keshav Santhanam, Andy Shih, Krishnan
  Srinivasan, Alex Tamkin, Rohan Taori, Armin~W. Thomas, Florian Tramèr,
  Rose~E. Wang, William Wang, Bohan Wu, Jiajun Wu, Yuhuai Wu, Sang~Michael Xie,
  Michihiro Yasunaga, Jiaxuan You, Matei Zaharia, Michael Zhang, Tianyi Zhang,
  Xikun Zhang, Yuhui Zhang, Lucia Zheng, Kaitlyn Zhou, and Percy Liang. 2021.
\newblock \href {http://arxiv.org/abs/2108.07258} {On the opportunities and
  risks of foundation models}.

\bibitem[{Brown et~al.(2020)Brown, Mann, Ryder, Subbiah, Kaplan, Dhariwal,
  Neelakantan, Shyam, Sastry, Askell, Agarwal, Herbert-Voss, Krueger, Henighan,
  Child, Ramesh, Ziegler, Wu, Winter, Hesse, Chen, Sigler, Litwin, Gray, Chess,
  Clark, Berner, McCandlish, Radford, Sutskever, and Amodei}]{brown2020gpt3}
Tom Brown, Benjamin Mann, Nick Ryder, Melanie Subbiah, Jared~D Kaplan, Prafulla
  Dhariwal, Arvind Neelakantan, Pranav Shyam, Girish Sastry, Amanda Askell,
  Sandhini Agarwal, Ariel Herbert-Voss, Gretchen Krueger, Tom Henighan, Rewon
  Child, Aditya Ramesh, Daniel Ziegler, Jeffrey Wu, Clemens Winter, Chris
  Hesse, Mark Chen, Eric Sigler, Mateusz Litwin, Scott Gray, Benjamin Chess,
  Jack Clark, Christopher Berner, Sam McCandlish, Alec Radford, Ilya Sutskever,
  and Dario Amodei. 2020.
\newblock \href
  {https://proceedings.neurips.cc/paper/2020/file/1457c0d6bfcb4967418bfb8ac142f64a-Paper.pdf}
  {Language models are few-shot learners}.
\newblock In \emph{Advances in Neural Information Processing Systems},
  volume~33, pages 1877--1901. Curran Associates, Inc.

\bibitem[{Chowdhery et~al.(2022)Chowdhery, Narang, Devlin, Bosma, Mishra,
  Roberts, Barham, Chung, Sutton, Gehrmann, Schuh, Shi, Tsvyashchenko, Maynez,
  Rao, Barnes, Tay, Shazeer, Prabhakaran, Reif, Du, Hutchinson, Pope, Bradbury,
  Austin, Isard, Gur-Ari, Yin, Duke, Levskaya, Ghemawat, Dev, Michalewski,
  Garcia, Misra, Robinson, Fedus, Zhou, Ippolito, Luan, Lim, Zoph, Spiridonov,
  Sepassi, Dohan, Agrawal, Omernick, Dai, Pillai, Pellat, Lewkowycz, Moreira,
  Child, Polozov, Lee, Zhou, Wang, Saeta, Diaz, Firat, Catasta, Wei,
  Meier-Hellstern, Eck, Dean, Petrov, and Fiedel}]{chowdhery2022palm}
Aakanksha Chowdhery, Sharan Narang, Jacob Devlin, Maarten Bosma, Gaurav Mishra,
  Adam Roberts, Paul Barham, Hyung~Won Chung, Charles Sutton, Sebastian
  Gehrmann, Parker Schuh, Kensen Shi, Sasha Tsvyashchenko, Joshua Maynez,
  Abhishek Rao, Parker Barnes, Yi~Tay, Noam Shazeer, Vinodkumar Prabhakaran,
  Emily Reif, Nan Du, Ben Hutchinson, Reiner Pope, James Bradbury, Jacob
  Austin, Michael Isard, Guy Gur-Ari, Pengcheng Yin, Toju Duke, Anselm
  Levskaya, Sanjay Ghemawat, Sunipa Dev, Henryk Michalewski, Xavier Garcia,
  Vedant Misra, Kevin Robinson, Liam Fedus, Denny Zhou, Daphne Ippolito, David
  Luan, Hyeontaek Lim, Barret Zoph, Alexander Spiridonov, Ryan Sepassi, David
  Dohan, Shivani Agrawal, Mark Omernick, Andrew~M. Dai,
  Thanumalayan~Sankaranarayana Pillai, Marie Pellat, Aitor Lewkowycz, Erica
  Moreira, Rewon Child, Oleksandr Polozov, Katherine Lee, Zongwei Zhou, Xuezhi
  Wang, Brennan Saeta, Mark Diaz, Orhan Firat, Michele Catasta, Jason Wei,
  Kathy Meier-Hellstern, Douglas Eck, Jeff Dean, Slav Petrov, and Noah Fiedel.
  2022.
\newblock \href {https://doi.org/10.48550/ARXIV.2204.02311} {Palm: Scaling
  language modeling with pathways}.

\bibitem[{Dagan et~al.(2006)Dagan, Glickman, and Magnini}]{dagan2006rte1}
Ido Dagan, Oren Glickman, and Bernardo Magnini. 2006.
\newblock The pascal recognising textual entailment challenge.
\newblock In \emph{Machine Learning Challenges. Evaluating Predictive
  Uncertainty, Visual Object Classification, and Recognising Tectual
  Entailment}, pages 177--190, Berlin, Heidelberg. Springer Berlin Heidelberg.

\bibitem[{Dai and Le(2015)}]{dai2015semisupervised}
Andrew~M Dai and Quoc~V Le. 2015.
\newblock \href
  {https://proceedings.neurips.cc/paper/2015/file/7137debd45ae4d0ab9aa953017286b20-Paper.pdf}
  {Semi-supervised sequence learning}.
\newblock In \emph{Advances in Neural Information Processing Systems},
  volume~28. Curran Associates, Inc.

\bibitem[{Devlin et~al.(2019)Devlin, Chang, Lee, and
  Toutanova}]{devlin-etal-2019-bert}
Jacob Devlin, Ming-Wei Chang, Kenton Lee, and Kristina Toutanova. 2019.
\newblock \href {https://doi.org/10.18653/v1/N19-1423} {{BERT}: Pre-training of
  deep bidirectional transformers for language understanding}.
\newblock In \emph{Proceedings of the 2019 Conference of the North {A}merican
  Chapter of the Association for Computational Linguistics: Human Language
  Technologies, Volume 1 (Long and Short Papers)}, pages 4171--4186,
  Minneapolis, Minnesota. Association for Computational Linguistics.

\bibitem[{Dolan and Brockett(2005)}]{dolan-brockett-2005-automatically}
William~B. Dolan and Chris Brockett. 2005.
\newblock \href {https://aclanthology.org/I05-5002} {Automatically constructing
  a corpus of sentential paraphrases}.
\newblock In \emph{Proceedings of the Third International Workshop on
  Paraphrasing ({IWP}2005)}.

\bibitem[{Giampiccolo et~al.(2007)Giampiccolo, Magnini, Dagan, and
  Dolan}]{giampiccolo-etal-2007-third}
Danilo Giampiccolo, Bernardo Magnini, Ido Dagan, and Bill Dolan. 2007.
\newblock \href {https://aclanthology.org/W07-1401} {The third {PASCAL}
  recognizing textual entailment challenge}.
\newblock In \emph{Proceedings of the {ACL}-{PASCAL} Workshop on Textual
  Entailment and Paraphrasing}, pages 1--9, Prague. Association for
  Computational Linguistics.

\bibitem[{Gokaslan and Cohen(2019)}]{gokaslan2019openwebtext}
Aaron Gokaslan and Vanya Cohen. 2019.
\newblock Openwebtext corpus.
\newblock
  \path{http://web.archive.org/save/http://Skylion007.github.io/OpenWebTextCorpus}.

\bibitem[{Haim et~al.(2006)Haim, Dagan, Dolan, Ferro, Giampiccolo, Magnini, and
  Szpektor}]{haim2006rte2}
R~Bar Haim, Ido Dagan, Bill Dolan, Lisa Ferro, Danilo Giampiccolo, Bernardo
  Magnini, and Idan Szpektor. 2006.
\newblock The second pascal recognising textual entailment challenge.
\newblock In \emph{Proceedings of the Second PASCAL Challenges Workshop on
  Recognising Textual Entailment}, volume~7.

\bibitem[{Hoffmann et~al.(2022)Hoffmann, Borgeaud, Mensch, Buchatskaya, Cai,
  Rutherford, Casas, Hendricks, Welbl, Clark, Hennigan, Noland, Millican,
  Driessche, Damoc, Guy, Osindero, Simonyan, Elsen, Rae, Vinyals, and
  Sifre}]{hoffmann2022chinchilla}
Jordan Hoffmann, Sebastian Borgeaud, Arthur Mensch, Elena Buchatskaya, Trevor
  Cai, Eliza Rutherford, Diego de~Las Casas, Lisa~Anne Hendricks, Johannes
  Welbl, Aidan Clark, Tom Hennigan, Eric Noland, Katie Millican, George van~den
  Driessche, Bogdan Damoc, Aurelia Guy, Simon Osindero, Karen Simonyan, Erich
  Elsen, Jack~W. Rae, Oriol Vinyals, and Laurent Sifre. 2022.
\newblock \href {https://doi.org/10.48550/ARXIV.2203.15556} {Training
  compute-optimal large language models}.

\bibitem[{Howard and Ruder(2018)}]{howard-ruder-2018-universal}
Jeremy Howard and Sebastian Ruder. 2018.
\newblock \href {https://doi.org/10.18653/v1/P18-1031} {Universal language
  model fine-tuning for text classification}.
\newblock In \emph{Proceedings of the 56th Annual Meeting of the Association
  for Computational Linguistics (Volume 1: Long Papers)}, pages 328--339,
  Melbourne, Australia. Association for Computational Linguistics.

\bibitem[{Joshi et~al.(2020)Joshi, Chen, Liu, Weld, Zettlemoyer, and
  Levy}]{joshi-etal-2020-spanbert}
Mandar Joshi, Danqi Chen, Yinhan Liu, Daniel~S. Weld, Luke Zettlemoyer, and
  Omer Levy. 2020.
\newblock \href {https://doi.org/10.1162/tacl_a_00300} {{S}pan{BERT}: Improving
  pre-training by representing and predicting spans}.
\newblock \emph{Transactions of the Association for Computational Linguistics},
  8:64--77.

\bibitem[{Lewis et~al.(2020)Lewis, Liu, Goyal, Ghazvininejad, Mohamed, Levy,
  Stoyanov, and Zettlemoyer}]{lewis-etal-2020-bart}
Mike Lewis, Yinhan Liu, Naman Goyal, Marjan Ghazvininejad, Abdelrahman Mohamed,
  Omer Levy, Veselin Stoyanov, and Luke Zettlemoyer. 2020.
\newblock \href {https://doi.org/10.18653/v1/2020.acl-main.703} {{BART}:
  Denoising sequence-to-sequence pre-training for natural language generation,
  translation, and comprehension}.
\newblock In \emph{Proceedings of the 58th Annual Meeting of the Association
  for Computational Linguistics}, pages 7871--7880, Online. Association for
  Computational Linguistics.

\bibitem[{Liu et~al.(2019)Liu, Ott, Goyal, Du, Joshi, Chen, Levy, Lewis,
  Zettlemoyer, and Stoyanov}]{liu2019roberta}
Yinhan Liu, Myle Ott, Naman Goyal, Jingfei Du, Mandar Joshi, Danqi Chen, Omer
  Levy, Mike Lewis, Luke Zettlemoyer, and Veselin Stoyanov. 2019.
\newblock \href {http://arxiv.org/abs/1907.11692} {Roberta: A robustly
  optimized bert pretraining approach}.

\bibitem[{Mihaylov et~al.(2018)Mihaylov, Clark, Khot, and
  Sabharwal}]{mihaylov-etal-2018-suit}
Todor Mihaylov, Peter Clark, Tushar Khot, and Ashish Sabharwal. 2018.
\newblock \href {https://doi.org/10.18653/v1/D18-1260} {Can a suit of armor
  conduct electricity? a new dataset for open book question answering}.
\newblock In \emph{Proceedings of the 2018 Conference on Empirical Methods in
  Natural Language Processing}, pages 2381--2391, Brussels, Belgium.
  Association for Computational Linguistics.

\bibitem[{Mostafazadeh et~al.(2016)Mostafazadeh, Chambers, He, Parikh, Batra,
  Vanderwende, Kohli, and Allen}]{mostafazadeh-etal-2016-corpus}
Nasrin Mostafazadeh, Nathanael Chambers, Xiaodong He, Devi Parikh, Dhruv Batra,
  Lucy Vanderwende, Pushmeet Kohli, and James Allen. 2016.
\newblock \href {https://doi.org/10.18653/v1/N16-1098} {A corpus and cloze
  evaluation for deeper understanding of commonsense stories}.
\newblock In \emph{Proceedings of the 2016 Conference of the North {A}merican
  Chapter of the Association for Computational Linguistics: Human Language
  Technologies}, pages 839--849, San Diego, California. Association for
  Computational Linguistics.

\bibitem[{Nagel(2016)}]{nagel2016ccnews}
Sebastian Nagel. 2016.
\newblock Cc-news.
\newblock
  \path{http://web.archive.org/save/http://commoncrawl.org/2016/10/news-dataset-available}.

\bibitem[{Ott et~al.(2019)Ott, Edunov, Baevski, Fan, Gross, Ng, Grangier, and
  Auli}]{ott-etal-2019-fairseq}
Myle Ott, Sergey Edunov, Alexei Baevski, Angela Fan, Sam Gross, Nathan Ng,
  David Grangier, and Michael Auli. 2019.
\newblock \href {https://doi.org/10.18653/v1/N19-4009} {fairseq: A fast,
  extensible toolkit for sequence modeling}.
\newblock In \emph{Proceedings of the 2019 Conference of the North {A}merican
  Chapter of the Association for Computational Linguistics (Demonstrations)},
  pages 48--53, Minneapolis, Minnesota. Association for Computational
  Linguistics.

\bibitem[{Peters et~al.(2018)Peters, Neumann, Iyyer, Gardner, Clark, Lee, and
  Zettlemoyer}]{peters-etal-2018-deep}
Matthew~E. Peters, Mark Neumann, Mohit Iyyer, Matt Gardner, Christopher Clark,
  Kenton Lee, and Luke Zettlemoyer. 2018.
\newblock \href {https://doi.org/10.18653/v1/N18-1202} {Deep contextualized
  word representations}.
\newblock In \emph{Proceedings of the 2018 Conference of the North {A}merican
  Chapter of the Association for Computational Linguistics: Human Language
  Technologies, Volume 1 (Long Papers)}, pages 2227--2237, New Orleans,
  Louisiana. Association for Computational Linguistics.

\bibitem[{Radford et~al.(2018)Radford, Narasimhan, Salimans, and
  Sutskever}]{radford2018gpt}
Alec Radford, Karthik Narasimhan, Time Salimans, and Ilya Sutskever. 2018.
\newblock Improving language understanding with unsupervised learning.
\newblock Technical report, OpenAI.

\bibitem[{Radford et~al.(2019)Radford, Wu, Child, Luan, Amodei, and
  Sutskever}]{radford2019gpt2}
Alec Radford, Jeffrey Wu, Rewon Child, David Luan, Dario Amodei, and Ilya
  Sutskever. 2019.
\newblock Language models are unsupervised multitask learners.
\newblock Technical report, OpenAI.

\bibitem[{Rae et~al.(2021)Rae, Borgeaud, Cai, Millican, Hoffmann, Song,
  Aslanides, Henderson, Ring, Young, Rutherford, Hennigan, Menick, Cassirer,
  Powell, Driessche, Hendricks, Rauh, Huang, Glaese, Welbl, Dathathri, Huang,
  Uesato, Mellor, Higgins, Creswell, McAleese, Wu, Elsen, Jayakumar,
  Buchatskaya, Budden, Sutherland, Simonyan, Paganini, Sifre, Martens, Li,
  Kuncoro, Nematzadeh, Gribovskaya, Donato, Lazaridou, Mensch, Lespiau,
  Tsimpoukelli, Grigorev, Fritz, Sottiaux, Pajarskas, Pohlen, Gong, Toyama,
  d'Autume, Li, Terzi, Mikulik, Babuschkin, Clark, Casas, Guy, Jones, Bradbury,
  Johnson, Hechtman, Weidinger, Gabriel, Isaac, Lockhart, Osindero, Rimell,
  Dyer, Vinyals, Ayoub, Stanway, Bennett, Hassabis, Kavukcuoglu, and
  Irving}]{rae2021gopher}
Jack~W. Rae, Sebastian Borgeaud, Trevor Cai, Katie Millican, Jordan Hoffmann,
  Francis Song, John Aslanides, Sarah Henderson, Roman Ring, Susannah Young,
  Eliza Rutherford, Tom Hennigan, Jacob Menick, Albin Cassirer, Richard Powell,
  George van~den Driessche, Lisa~Anne Hendricks, Maribeth Rauh, Po-Sen Huang,
  Amelia Glaese, Johannes Welbl, Sumanth Dathathri, Saffron Huang, Jonathan
  Uesato, John Mellor, Irina Higgins, Antonia Creswell, Nat McAleese, Amy Wu,
  Erich Elsen, Siddhant Jayakumar, Elena Buchatskaya, David Budden, Esme
  Sutherland, Karen Simonyan, Michela Paganini, Laurent Sifre, Lena Martens,
  Xiang~Lorraine Li, Adhiguna Kuncoro, Aida Nematzadeh, Elena Gribovskaya,
  Domenic Donato, Angeliki Lazaridou, Arthur Mensch, Jean-Baptiste Lespiau,
  Maria Tsimpoukelli, Nikolai Grigorev, Doug Fritz, Thibault Sottiaux, Mantas
  Pajarskas, Toby Pohlen, Zhitao Gong, Daniel Toyama, Cyprien de~Masson
  d'Autume, Yujia Li, Tayfun Terzi, Vladimir Mikulik, Igor Babuschkin, Aidan
  Clark, Diego de~Las Casas, Aurelia Guy, Chris Jones, James Bradbury, Matthew
  Johnson, Blake Hechtman, Laura Weidinger, Iason Gabriel, William Isaac,
  Ed~Lockhart, Simon Osindero, Laura Rimell, Chris Dyer, Oriol Vinyals, Kareem
  Ayoub, Jeff Stanway, Lorrayne Bennett, Demis Hassabis, Koray Kavukcuoglu, and
  Geoffrey Irving. 2021.
\newblock \href {https://doi.org/10.48550/ARXIV.2112.11446} {Scaling language
  models: Methods, analysis \& insights from training gopher}.

\bibitem[{Raffel et~al.(2020)Raffel, Shazeer, Roberts, Lee, Narang, Matena,
  Zhou, Li, and Liu}]{raffel2020t5}
Colin Raffel, Noam Shazeer, Adam Roberts, Katherine Lee, Sharan Narang, Michael
  Matena, Yanqi Zhou, Wei Li, and Peter~J. Liu. 2020.
\newblock \href {http://jmlr.org/papers/v21/20-074.html} {Exploring the limits
  of transfer learning with a unified text-to-text transformer}.
\newblock \emph{Journal of Machine Learning Research}, 21(140):1--67.

\bibitem[{Rajpurkar et~al.(2016)Rajpurkar, Zhang, Lopyrev, and
  Liang}]{rajpurkar-etal-2016-squad}
Pranav Rajpurkar, Jian Zhang, Konstantin Lopyrev, and Percy Liang. 2016.
\newblock \href {https://doi.org/10.18653/v1/D16-1264} {{SQ}u{AD}: 100,000+
  questions for machine comprehension of text}.
\newblock In \emph{Proceedings of the 2016 Conference on Empirical Methods in
  Natural Language Processing}, pages 2383--2392, Austin, Texas. Association
  for Computational Linguistics.

\bibitem[{Sakaguchi et~al.(2020)Sakaguchi, Le~Bras, Bhagavatula, and
  Choi}]{sakaguchi2020winogrande}
Keisuke Sakaguchi, Ronan Le~Bras, Chandra Bhagavatula, and Yejin Choi. 2020.
\newblock \href {https://doi.org/10.1609/aaai.v34i05.6399} {Winogrande: An
  adversarial winograd schema challenge at scale}.
\newblock \emph{Proceedings of the AAAI Conference on Artificial Intelligence},
  34(05):8732--8740.

\bibitem[{Smith et~al.(2022)Smith, Patwary, Norick, LeGresley, Rajbhandari,
  Casper, Liu, Prabhumoye, Zerveas, Korthikanti, Zhang, Child, Aminabadi,
  Bernauer, Song, Shoeybi, He, Houston, Tiwary, and Catanzaro}]{smith2022mtnlg}
Shaden Smith, Mostofa Patwary, Brandon Norick, Patrick LeGresley, Samyam
  Rajbhandari, Jared Casper, Zhun Liu, Shrimai Prabhumoye, George Zerveas,
  Vijay Korthikanti, Elton Zhang, Rewon Child, Reza~Yazdani Aminabadi, Julie
  Bernauer, Xia Song, Mohammad Shoeybi, Yuxiong He, Michael Houston, Saurabh
  Tiwary, and Bryan Catanzaro. 2022.
\newblock \href {https://doi.org/10.48550/ARXIV.2201.11990} {Using deepspeed
  and megatron to train megatron-turing nlg 530b, a large-scale generative
  language model}.

\bibitem[{Socher et~al.(2013)Socher, Perelygin, Wu, Chuang, Manning, Ng, and
  Potts}]{socher-etal-2013-recursive}
Richard Socher, Alex Perelygin, Jean Wu, Jason Chuang, Christopher~D. Manning,
  Andrew Ng, and Christopher Potts. 2013.
\newblock \href {https://aclanthology.org/D13-1170} {Recursive deep models for
  semantic compositionality over a sentiment treebank}.
\newblock In \emph{Proceedings of the 2013 Conference on Empirical Methods in
  Natural Language Processing}, pages 1631--1642, Seattle, Washington, USA.
  Association for Computational Linguistics.

\bibitem[{Sun et~al.(2021)Sun, Wang, Feng, Ding, Pang, Shang, Liu, Chen, Zhao,
  Lu, Liu, Wu, Gong, Liang, Shang, Sun, Liu, Ouyang, Yu, Tian, Wu, and
  Wang}]{sun2021ernie3}
Yu~Sun, Shuohuan Wang, Shikun Feng, Siyu Ding, Chao Pang, Junyuan Shang,
  Jiaxiang Liu, Xuyi Chen, Yanbin Zhao, Yuxiang Lu, Weixin Liu, Zhihua Wu,
  Weibao Gong, Jianzhong Liang, Zhizhou Shang, Peng Sun, Wei Liu, Xuan Ouyang,
  Dianhai Yu, Hao Tian, Hua Wu, and Haifeng Wang. 2021.
\newblock \href {https://doi.org/10.48550/ARXIV.2107.02137} {Ernie 3.0:
  Large-scale knowledge enhanced pre-training for language understanding and
  generation}.

\bibitem[{Tay et~al.(2022{\natexlab{a}})Tay, Dehghani, Abnar, Chung, Fedus,
  Rao, Narang, Tran, Yogatama, and Metzler}]{anonymous2021scaling}
Yi~Tay, Mostafa Dehghani, Samira Abnar, Hyung~Won Chung, William Fedus, Jinfeng
  Rao, Sharan Narang, Vinh~Q. Tran, Dani Yogatama, and Donald Metzler.
  2022{\natexlab{a}}.
\newblock \href {https://doi.org/10.48550/ARXIV.2207.10551} {Scaling laws vs
  model architectures: How does inductive bias influence scaling?}

\bibitem[{Tay et~al.(2022{\natexlab{b}})Tay, Dehghani, Rao, Fedus, Abnar,
  Chung, Narang, Yogatama, Vaswani, and Metzler}]{tay2022scale}
Yi~Tay, Mostafa Dehghani, Jinfeng Rao, William Fedus, Samira Abnar, Hyung~Won
  Chung, Sharan Narang, Dani Yogatama, Ashish Vaswani, and Donald Metzler.
  2022{\natexlab{b}}.
\newblock \href {https://openreview.net/forum?id=f2OYVDyfIB} {Scale
  efficiently: Insights from pretraining and finetuning transformers}.
\newblock In \emph{International Conference on Learning Representations}.

\bibitem[{Trinh and Le(2018)}]{trinh2018simple}
Trieu~H. Trinh and Quoc~V. Le. 2018.
\newblock \href {https://doi.org/10.48550/ARXIV.1806.02847} {A simple method
  for commonsense reasoning}.

\bibitem[{Wang et~al.(2019)Wang, Singh, Michael, Hill, Levy, and
  Bowman}]{wang2018glue}
Alex Wang, Amanpreet Singh, Julian Michael, Felix Hill, Omer Levy, and
  Samuel~R. Bowman. 2019.
\newblock \href {https://openreview.net/forum?id=rJ4km2R5t7} {{GLUE}: A
  multi-task benchmark and analysis platform for natural language
  understanding}.
\newblock In \emph{International Conference on Learning Representations}.

\bibitem[{Wang et~al.(2022)Wang, Roberts, Hesslow, Scao, Chung, Beltagy,
  Launay, and Raffel}]{wang2022language}
Thomas Wang, Adam Roberts, Daniel Hesslow, Teven~Le Scao, Hyung~Won Chung,
  Iz~Beltagy, Julien Launay, and Colin Raffel. 2022.
\newblock \href {http://arxiv.org/abs/2204.05832} {What language model
  architecture and pretraining objective work best for zero-shot
  generalization?}

\bibitem[{Warstadt et~al.(2019)Warstadt, Singh, and
  Bowman}]{warstadt-etal-2019-neural}
Alex Warstadt, Amanpreet Singh, and Samuel~R. Bowman. 2019.
\newblock \href {https://doi.org/10.1162/tacl_a_00290} {Neural network
  acceptability judgments}.
\newblock \emph{Transactions of the Association for Computational Linguistics},
  7:625--641.

\bibitem[{Wenzek et~al.(2020)Wenzek, Lachaux, Conneau, Chaudhary, Guzm{\'a}n,
  Joulin, and Grave}]{wenzek-etal-2020-ccnet}
Guillaume Wenzek, Marie-Anne Lachaux, Alexis Conneau, Vishrav Chaudhary,
  Francisco Guzm{\'a}n, Armand Joulin, and Edouard Grave. 2020.
\newblock \href {https://aclanthology.org/2020.lrec-1.494} {{CCN}et: Extracting
  high quality monolingual datasets from web crawl data}.
\newblock In \emph{Proceedings of the 12th Language Resources and Evaluation
  Conference}, pages 4003--4012, Marseille, France. European Language Resources
  Association.

\bibitem[{Williams et~al.(2018)Williams, Nangia, and
  Bowman}]{williams-etal-2018-broad}
Adina Williams, Nikita Nangia, and Samuel Bowman. 2018.
\newblock \href {https://doi.org/10.18653/v1/N18-1101} {A broad-coverage
  challenge corpus for sentence understanding through inference}.
\newblock In \emph{Proceedings of the 2018 Conference of the North {A}merican
  Chapter of the Association for Computational Linguistics: Human Language
  Technologies, Volume 1 (Long Papers)}, pages 1112--1122, New Orleans,
  Louisiana. Association for Computational Linguistics.

\bibitem[{Wu et~al.(2021)Wu, Zhao, Yu, Zhang, Shen, Liu, Li, Zhu, Luo, Xu, and
  Zhang}]{wu2021yuan}
Shaohua Wu, Xudong Zhao, Tong Yu, Rongguo Zhang, Chong Shen, Hongli Liu, Feng
  Li, Hong Zhu, Jiangang Luo, Liang Xu, and Xuanwei Zhang. 2021.
\newblock \href {https://doi.org/10.48550/ARXIV.2110.04725} {Yuan 1.0:
  Large-scale pre-trained language model in zero-shot and few-shot learning}.

\bibitem[{Yang et~al.(2019)Yang, Dai, Yang, Carbonell, Salakhutdinov, and
  Le}]{yang2019xlnet}
Zhilin Yang, Zihang Dai, Yiming Yang, Jaime Carbonell, Russ~R Salakhutdinov,
  and Quoc~V Le. 2019.
\newblock \href
  {https://proceedings.neurips.cc/paper/2019/file/dc6a7e655d7e5840e66733e9ee67cc69-Paper.pdf}
  {Xlnet: Generalized autoregressive pretraining for language understanding}.
\newblock In \emph{Advances in Neural Information Processing Systems},
  volume~32. Curran Associates, Inc.

\bibitem[{Zellers et~al.(2019)Zellers, Holtzman, Bisk, Farhadi, and
  Choi}]{zellers-etal-2019-hellaswag}
Rowan Zellers, Ari Holtzman, Yonatan Bisk, Ali Farhadi, and Yejin Choi. 2019.
\newblock \href {https://doi.org/10.18653/v1/P19-1472} {{H}ella{S}wag: Can a
  machine really finish your sentence?}
\newblock In \emph{Proceedings of the 57th Annual Meeting of the Association
  for Computational Linguistics}, pages 4791--4800, Florence, Italy.
  Association for Computational Linguistics.

\bibitem[{Zhang et~al.(2018)Zhang, Liu, Liu, Gao, Duh, and
  Van~Durme}]{zhang2018record}
Sheng Zhang, Xiaodong Liu, Jingjing Liu, Jianfeng Gao, Kevin Duh, and Benjamin
  Van~Durme. 2018.
\newblock \href {https://doi.org/10.48550/ARXIV.1810.12885} {Record: Bridging
  the gap between human and machine commonsense reading comprehension}.

\bibitem[{Zhang et~al.(2022)Zhang, Roller, Goyal, Artetxe, Chen, Chen, Dewan,
  Diab, Li, Lin, Mihaylov, Ott, Shleifer, Shuster, Simig, Koura, Sridhar, Wang,
  and Zettlemoyer}]{zhang2022opt}
Susan Zhang, Stephen Roller, Naman Goyal, Mikel Artetxe, Moya Chen, Shuohui
  Chen, Christopher Dewan, Mona Diab, Xian Li, Xi~Victoria Lin, Todor Mihaylov,
  Myle Ott, Sam Shleifer, Kurt Shuster, Daniel Simig, Punit~Singh Koura, Anjali
  Sridhar, Tianlu Wang, and Luke Zettlemoyer. 2022.
\newblock \href {https://doi.org/10.48550/ARXIV.2205.01068} {{OPT}: Open
  pre-trained transformer language models}.

\bibitem[{Zhu et~al.(2015)Zhu, Kiros, Zemel, Salakhutdinov, Urtasun, Torralba,
  and Fidler}]{zhu2015bookcorpus}
Yukun Zhu, Ryan Kiros, Richard Zemel, Ruslan Salakhutdinov, Raquel Urtasun,
  Antonio Torralba, and Sanja Fidler. 2015.
\newblock \href {https://doi.org/10.48550/ARXIV.1506.06724} {Aligning books and
  movies: Towards story-like visual explanations by watching movies and reading
  books}.

\end{thebibliography}
\bibliographystyle{acl_natbib}

\appendix

\section{Proposed framework}
\label{app:framework}

\begin{figure*}[t]
\centering
\includegraphics[width=0.8\linewidth]{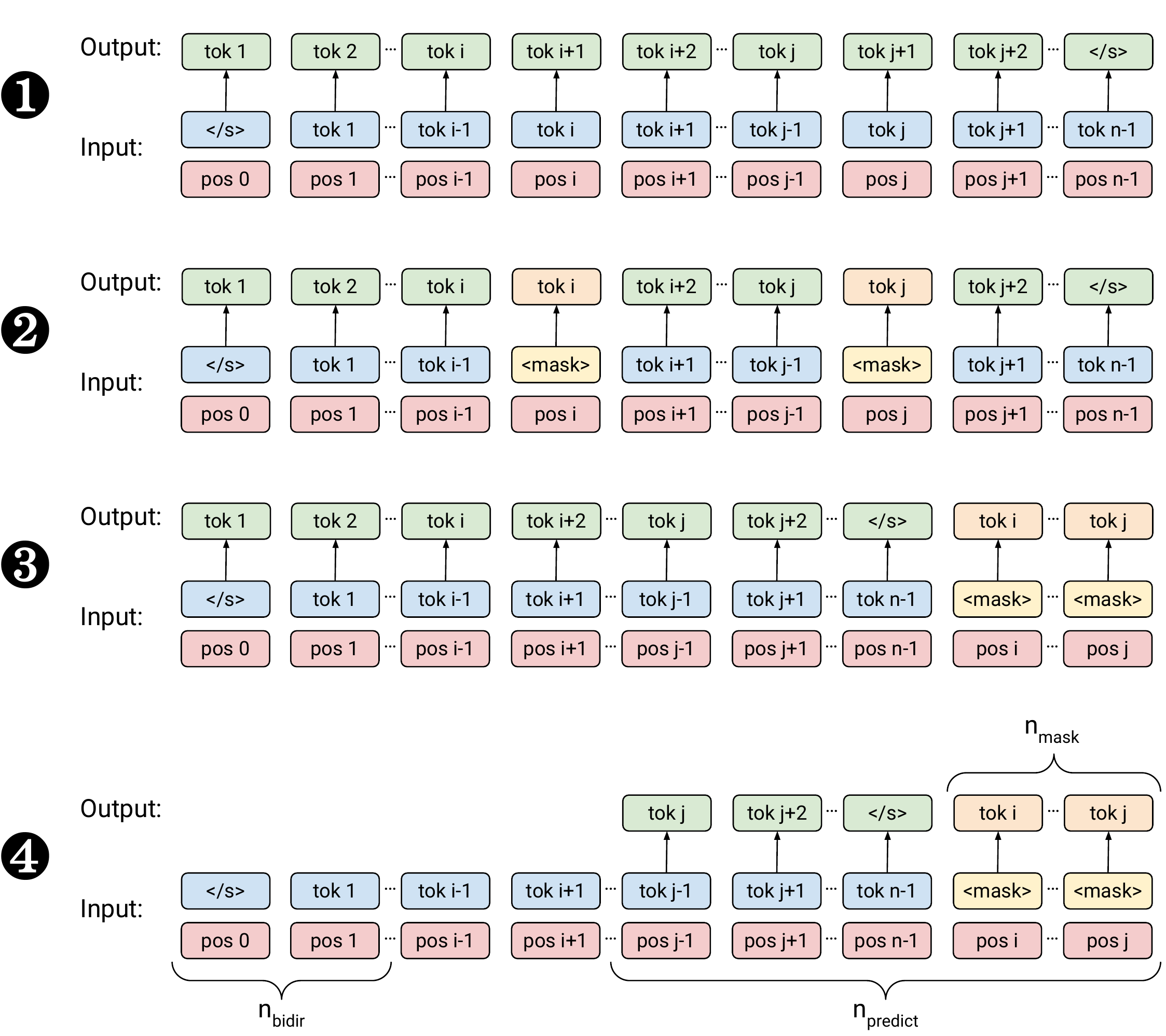}
\caption{
\textbf{Proposed framework.}
1) We start with the original sequence in the input, and predict the next token in the output; 2) We choose $\nmask$ tokens at random, replace them with the special <mask> token in the input, and predict the masked token (rather than the next token) in the output; 3) We move the masked tokens and their corresponding positional embeddings to the end; 4) We only predict the last $\npredict$ tokens, using bidirectional attention for the first $\nbidir$ tokens and unidirectional attention for the rest (final objective).
}
\label{fig:framework_stepwise}
\end{figure*}

Figure \ref{fig:framework_stepwise} provides a step-by-step description of how we define our objective starting from the original sequence.

\section{Fine-tuning settings}
\label{app:finetuning}

For fine-tuning, we did grid search on learning rate $\in \{5e-06, 5e-05, 1e-05, 2e-05\}$ and batch size $\in \{16, 32, 64\}$.
For each task, we trained the same numbers of updates for different setups and reported the best numbers across the grid.
The details of fine-tuning tasks and numbers of updates can be found in Table \ref{tab:ft_tasks}, which were chosen to follow the original settings from RoBERTa.
We used Adam and polynomial decay scheduler for optimization.

\begin{table}[t]
\begin{center}
\begin{small}
\addtolength{\tabcolsep}{-2.5pt}
\begin{tabular}{rr}
\toprule
\multicolumn{1}{c}{\emph{task}} & \multicolumn{1}{c}{\emph{\# of updates}} \\
\midrule
CoLA & 5336
\\
SST-2 & 20935
\\
MNLI & 123873
\\
QNLI & 33112
\\
MRPC & 2296
\\
RTE & 2036
\\
\bottomrule
\end{tabular}%
\end{small}
\end{center}
\caption{Number of fine-tuning updates for each task.}
\label{tab:ft_tasks}
\end{table}

\end{document}